\documentclass{article}

\usepackage{amsmath}
\usepackage{graphicx}
\usepackage[font=small,labelfont=bf]{caption}
\usepackage{amsfonts}
\usepackage{amsthm}
\usepackage{thmtools}
\usepackage{amssymb}
\usepackage{xcolor}
\usepackage{algpseudocode}

\title{Hierarchy Representation of Data in Machine Learnings}
\author{Han Yegang \\My Paul School \\rornfl9909@naver.com \and Park Minjun \\ My Paul School \\happyjune1022@naver.com \and Byun Duwon\\My Paul School\\math2061@gmail.com
\and Park Inkyu \\ 
Department of Electronics \\
Information and Communication Engineering \\
Kangwon National University \\ eleriron@kangwon.ac.kr}
\date{\today}

\begin{document}

\maketitle

\begin{abstract}

When there are models with clear-cut judgment results for several data points, it is possible that most models exhibit a relationship where if they correctly judge one target, they also correctly judge another target. Conversely, if most models incorrectly judge one target, they may also incorrectly judge another target. We propose a method for visualizing this hierarchy among targets. This information is expected to be beneficial for model improvement.

\end{abstract}

%%%%%%%%%% macro definition %%%%%%%%%%%%%
\def\cal#1{\mathcal#1}
\def\b#1{\Bbb#1}
\def\k{\cal{K}}
%%%%%%%%%%%%%%%%%%%%%%%%%%%%%%%%%%%%%%%%%

\setcounter{secnumdepth}{3}

\nocite{11,12,13,14,15,16,17,18,19,20}

\section{Introduction}
Machine learning is currently the subject of extensive research efforts aimed at enhancing its performance. Research that exclusively emphasizes an empiricist perspective, viewing all knowledge as derived from empirical experiences, has enjoyed a great success. However, for effective model improvement, it is essential not only to focus on the model but also to investigate the data. Balancing the rational perspective, which estimates models based on datasets, is essential for improving the model's learning process([8]). Additionally, it is argued that understanding the cause-and-effect perspective, independent of how data fits, is vital, and acquiring such a perspective is crucial([9]). In this paper, we address another important aspect that understands and interprets data through the relationships within data. Based on the theoretical analysis, we investigate the hierarchy of data for a series of models to find data challenges in improving the models.

We examine the hierarchy of data using the concept of knowledge spaces theory. Knowledge spaces theory analyzes the knowledge structure at each stage based on students' learning evaluation results, aiming to enhance students' learning efficiency. In Subsection 2.3, we provide a visual representation of the hierarchy based on practical examples. This approach underscores its applicability not limited to simple examples but also extends to more complex machine learning environments. As a result, we provide a model monitoring approach and conduct information analysis for model improvement.

\section{Extending and applying knowledge space theory}
\subsection{Overview of knowledge space theory}
\subsubsection{knowledge structure}
In the acquisition of certain knowledge, there is often a need for a specific sequence or hierarchy. This hierarchy is referred to as the hierarchy of learning, and it is well-reflected in the field of testing. Of course, the following  two assumptions are necessary.
\begin{itemize}
    \item There are no accidental correct answers to unfamiliar questions.
    \item There are no accidental incorrect answers to questions that can be answered correctly.
\end{itemize}

Inversely, under the assumptions mentioned above, if testing is conducted on enough students, it is possible to investigate the hierarchy of knowledge. The theory used for this purpose is knowledge space theory(\cite{11}).

The set of assessment items is denoted as $Q$, and the set of questions correctly answered by a student is referred to as their knowledge state. Let $\k$ be a set of knowledge states. In this case, if the set $\k$ includes both the empty set $\phi$ and the entire set $Q$, then the ordered pair $(Q,\k)$ is referred to as a knowledge structure.

If $(Q,\k)$ is a knowledge structure, then the union of elements in $\k$ is equal to $Q$, i.e., $\bigcup_{K \in \k}K = Q$ holds. Therefore, unless there is any specific confusion, a knowledge structure $(Q,\k)$ is denoted simply by $\k$.

A knowledge structure is a set obtained from the assessment results. If this set lacks either $\phi$ or $Q$, it is possible to add them to construct the knowledge structure. The empty set represents a student who answered all questions incorrectly, while the entire set $Q$ signifies the existence of a student who answered all questions correctly. Therefore, we consider each element of the knowledge structure $\k$ as a state of knowledge. This approach is a valid definition when considering real-world scenarios.

\subsubsection{hierarchy}
We define symbols. For a knowledge structure $(Q,\k)$ and an element $q$ of $Q$, we define $\k_q$ as the set of all knowledge states that include $q$. i.e., $\k_q =\{K \in \k \,\, | \,\, q \in K \}$.

For example, we can illustrate this with a set of assessment items $Q=\{a,b,c,d,e\}$. We let
$$
\k=\{\phi, \{b,c\}, \{a,b,c\}, \{a,b,c,d\}, \{a,b,c,e\}, Q\}
$$ 

In this case, $(Q,\k)$ forms a knowledge structure. It can be represented as follows:

\begin{align*}
\k_a&=\{\{a,b,c\}, \{a,b,c,d\}, \{a,b,c,e\}, Q \} \\
\k_b&=\{\{b,c\}, \{a,b,c\}, \{a,b,c,d\}, \{a,b,c,e\},Q \} \\
\k_c&=\{\{b,c\}, \{a,b,c\}, \{a,b,c,d\}, \{a,b,c,e\},Q \} \\
\k_d&=\{\{a,b,c,d\}, Q\} \\
\k_e&=\{\{a,b,c,e\}, Q\}
\end{align*}

In particular $\k_b = \k_c$ holds. This implies that any knowledge state containing item $b$ also contains item $c$, and vice versa. In other words, all students who answer item $b$ correctly can also answer item $c$ correctly, and vice versa. Therefore, intuitively, items $b,c$ provide us with the same information. This is referred to as items $b$ and $c$ being equally informative.

We define the following for a knowledge structure $(Q,\k)$ and two elements $p,q$ from $Q$:

$$
p \rightarrow q  \Longleftrightarrow p \in \bigcap \k_q
$$

$ u \Longleftrightarrow v$ means the definition of $u$ as $v$. We refer to the relation $\rightarrow$ as the surmise relation, which becomes a quasi-order relation on $Q$(\cite{11}, p.36).

[Figure 1] illustrates our example using arrows.

\begin{center}
\includegraphics[width=1.1cm, height=1.7cm]{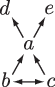}
\captionof{figure}{hierarchy of learning}
\end{center}

\subsubsection{pre-order and order relationship on knowledge space}
For a knowledge structure $(Q,\k)$, when the elements $p,q$ of $Q$ share the same information source, i.e., $p \leftrightarrow q$, we define this as the relation $p\sim q$. It can be easily verified that $(Q,\sim)$ forms an equivalence relation. Using this equivalence relation, we can partition the set $Q$ as follows within the previously defined knowledge structure $(Q,\k)$(\cite{12}).

\begin{center}
\includegraphics[width=4cm, height=2cm]{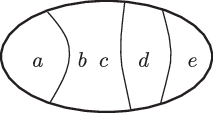}
\end{center}

These equivalence classes determined by the equivalence relation are referred to as concepts, denoted by $*$,  In our example, we have $b^* = c^*$, and $a^* , b^*, d^* , e^* $ are concepts

For a knowledge structure $(Q,\k)$, when every concept consists of only one element, i.e., for every element $q$, $q^*$ is composed of only one element, the knowledge structure $(Q,\k)$ is referred to as discriminative. In particular, if the knowledge structure $(Q,\k)$ is discriminative, the relation $\rightarrow$ becomes an order relation(\cite{11}, p.36).

When a knowledge structure $(Q,\k)$ is discriminative, it holds true that $q^* =\{q\}$ for every element $q$ in $Q$. Therefore, every element in $Q$ belongs to distinct concepts.

To facilitate understanding, we explain the previously defined knowledge structure $(Q,\k)$. For a subset $A$ of $Q$, we use the notation $A^* =\{q^* \,\, |\,\, q \in A \}$. Applying this notation, we have
$$
Q^*=\{a^* , b^* , d^*, e^* \}
$$

The elements within $Q^*$ represent items that belong to the same concept, grouped together. Regarding the knowledge structure $\k$, we define $\k^*$ as follows:

$\k^*=\{K^* \,\, |\,\, K \in \k \}$

Thus, $(Q^*,\k^*)$ forms a single knowledge structure, and it is discriminative. We refer to this as the discriminative reduction of the knowledge structure $(Q,\k)$. When we compute the discriminative reduction for the previously defined knowledge structure $(Q,\k)$, we get
$$
\k^*=\{\phi, \{b^*\}, \{a^*, b^* \}, \{a^*, b^*, d^*\}, \{a^*, b^* , e^* \}, Q^* \}
$$

The above method can be applied to transform the given knowledge structure into a discriminative knowledge structure, as explained earlier.

\subsection{Applying to machine learning}
\subsubsection{extension to machine learning}
We assume the development of AI machines for tasks like image recognition and natural language processing, where the AI systems can make clear correct or incorrect answers. We can apply the content of previous section to this scenario by treating each machine undergoing training as a student and considering the test questions as the objects of assessment. This scenario is more straightforward than the previous one. In the previous section, we could discuss the theory under two basic assumptions because it was focused on interactions with humans. However, in the case of machines, there is no need for such assumptions. Unless there is confusion, we can use the symbols from the previous section without any changes.

We denote the objects of judgement as the set $Q$, which can be represented as follows:

$$
Q=\{t_1, t_2, \cdots, t_l \}
$$

For the set of targets $Q$, we assume a series of learning machines $M=\{M_1, M_2, \cdots, M_m\}$ .

We consider the machine $M_i$ judgment of target $t_j$, representing correctness as $1$ and incorrectness as $0$ in the form of a function.

$$
M_i(t_j)=
\begin{cases}
1 & \text{(if the judgment is correct)} \\
0 & \text{(if the judgment is incorrect)}
\end{cases}
$$

In this case, we can view $\k$ as the set of knowledge states defined as:
$$
\k=\{ S \subset Q \,|\, S=M_i^{-1}(\{1\}) \quad i=1,2, \cdots, m\}
$$
where $S$ represents the set of correctly judged targets for each machine $M_i$.

For $p, q \in Q$, we assume that there is no machine that judges $p$ incorrectly and $q$ correctly. This condition can be expressed as $\k_p^c \cap \k_q = \phi$, equivalently $\k_q - \k_p = \phi$. Furthermore, this implies $\k_q \subset \k_p $, and once again, this means $p \in \cap \k_q$, leading to the conclusion that $p\rightarrow q$.

Furthermore, for $p,q \in Q$, if $p$ and $q$ are equally informative,
$$
M_i (p) = M_i (q) \quad i=1,2, \cdots, m
$$
This can be easily verified. In the knowledge structure $\k$, if $p^* = q^*$, then $\k_p =\k_q$, and as a result, the above equation holds.

In the preceding section concerning knowledge space theory, we delved into various sets to establish order relationships. Nevertheless, it becomes evident that we can also derive order relationships by considering the absence of a machines that judges one object incorrectly while judging another correctly. This methodology will be applied in Subsection 2.3.

\subsubsection{flexibility of judgements}
When defining the previous order relationships, the condition of having no machines that satisfy the criteria, in other words, a count of 0, played a crucial role. However, this concept can be extended to a probabilistic context. Instead of precisely 0, we can treat values close to 0 similarly. This method is elaborated on in \cite{14}, but since it's written in Korean, we provide a brief description here.

When introducing a flexibility of $m$\% ($m < 50$) for the relationship $\hookrightarrow$ in the context of Table 1, we define $p\hookrightarrow q$ as follows:

\begin{equation}\label{ordering}
p=q \quad \text{or} \quad \frac{n_3}{n_2 +n_3 } \times 100  \leq m  \tag{*}
\end{equation}

\begin{table}[ht]
\centering
\caption{judgments for $p,q$}
\begin{center}
\begin{tabular}[centerline]{|c|c|c|} \hline
$p$& $q$ & count  \\  \hline
1  & 1   &  $n_1$ \\ \hline
1  & 0   &  $n_2$ \\ \hline
0  & 1   &  $n_3$ \\ \hline
0  & 0   &  $n_4$ \\ \hline
\end{tabular}
\end{center}
\end{table}

The meaning of $p \hookrightarrow q$ is that most machines that answer $q$ correctly also answer $p$ correctly.'
It's clear that the relation $\hookrightarrow$ is reflexive. If we assume $p \hookrightarrow q$ and $q \hookrightarrow p$, then it implies $p = q$. If not, then we have
$$
\frac{n_3}{n_2 + n_3 } \times 100 \leq m \quad \text{and} \quad \frac{n_2}{n_2 +n_3 }\times 100 \leq m
$$
Adding these inequalities leads to $m \geq 50$, which is not possible. Therefore, the relation $\hookrightarrow$ is anti-symmetric.

We prove that the relation $\hookrightarrow$ is transitive. Assume $p \hookrightarrow q$ and $q \hookrightarrow r$ for objects $p, q, r$. Then, the answers of machines for these objects are as shown in the following table:

\begin{table}[ht]
\centering
\caption{judgments for $p,q,r$}
\begin{center}
\begin{tabular}[centerline]{|c|c|c|c|} \hline
$p$& $q$ & $r$ & count  \\  \hline
0  & 1   &1   &  $m_1$ \\ \hline
1  & 0   &1   &  $m_2$ \\ \hline
1  & 1   &0   &  $m_3$ \\ \hline
0  & 0   &1   &  $m_4$ \\ \hline
1  & 0   &0   &  $m_5$ \\ \hline
0  & 1   &0  &  $m_6$ \\ \hline
\end{tabular}
\end{center}
\end{table}

Since $p \hookrightarrow q$ and $q \hookrightarrow p$, we have
\begin{align*}
&\frac{m_1 +m_6}{m_1 + m_2 + m_5 + m_6} \times 100 \leq m, \\
&\frac{m_2 +m_4}{m_2 + m_3 + m_4 + m_6} \times 100 \leq m 
\end{align*}

Rearranging these two inequalities, we obtain
\begin{align*}
&100(m_{1} +m_{6}) \leq m(m_{1} +m_{2} +m_{5} +m_{6} ) \\
&100(m_{2} +m_{4}) \leq m(m_{2} +m_{3} +m_{4} +m_{6} )
\end{align*}

By adding the left-hand sides together and the right-hand sides together, we get
$$
100(m_{1} +m_{4}) \leq m(m_{1} +m_{3} +m_{4} +m_{5} )+2(m-50)(m_{2} +m_{6} )
$$
Furthermore, since $m < 50$, we have
$$
100(m_{1} +m_{4}) < m(m_{1} +m_{3} +m_{4} +m_{5} )
$$
This implies $p \rightarrow r$. Thus, we can conclude that the relation $\hookrightarrow$ is indeed an order relation([2], p.73).
From now on, we will use $\rightarrow$ instead of $\hookrightarrow$. Additionally, when $p$ and $q$ are in the upper and lower positions, respectively, we will represent $p \hookrightarrow q$ or $p \rightarrow q$ as \,\,
$\begin{matrix}q\\|\\p\end{matrix}$\,\,

\subsection{experiment}

In this subsection, we explain the practical application of knowledge space theory. 
To focus on the application method, we use 10 targets and 12 models.
The data here is arbitary.

Typically, when we train a model, we go through a process of improving it by randomly setting initial parameters. Depending on these parameter settings, models can either improve or degrade in performance, even when trained on the same data. 
A series of models yields defferent results from the same dataset.
We are curious about which data poses challenges for the models within this series, and we present a method to visualize this information

\begin{table}[hbt!]
\centering
\caption{judgments of models}
\begin{center}
\begin{tabular}[centerline]{|c|c|c|c|c|c|c|c|c|c|c|} \hline
       & $t_0$ & $t_1$ & $t_2$ & $t_3$ & $t_4$ & $t_5$ & $t_6$ & $t_7$ & $t_8$ &$t_9$   \\  \hline
 $M_{1}$ &1 & 1 & 0 & 0 & 0 & 0 & 0 & 0 & 0 & 0 \\ \hline
 $M_{2}$ &1 & 1 & 0 & 0 & 1 & 0 & 0 & 0 & 0 & 0 \\ \hline
 $M_{3}$ &1 & 1 & 1 & 0 & 1 & 0 & 1 & 0 & 0 & 0 \\ \hline
 $M_{4}$ &1 & 1 & 1 & 0 & 1 & 1 & 1 & 0 & 0 & 0 \\ \hline
 $M_{5}$ &1 & 1 & 1 & 0 & 1 & 1 & 1 & 0 & 0 & 1 \\ \hline
 $M_{6}$ &1 & 1 & 1 & 1 & 1 & 1 & 1 & 0 & 0 & 1 \\ \hline
 $M_{7}$ &1 & 1 & 0 & 0 & 0 & 1 & 1 & 0 & 0 & 0 \\ \hline
 $M_{8}$ &1 & 1 & 1 & 0 & 0 & 1 & 1 & 0 & 0 & 0 \\ \hline
 $M_{9}$ &1 & 1 & 1 & 1 & 0 & 1 & 1 & 0 & 0 & 0 \\ \hline
 $M_{10}$ &1 & 1 & 1 & 1 & 1 & 1 & 1 & 0 & 0 & 0 \\ \hline
 $M_{11}$ &1 & 1 & 1 & 1 & 0 & 1 & 1 & 0 & 1 & 0 \\ \hline
 $M_{12}$ &1 & 1 & 1 & 1 & 1 & 1 & 1 & 1 & 1 & 0 \\ \hline
\end{tabular}
\end{center}
\end{table}
Using Table 4, we can visualize the hierarchy of targets.

Figure 2 represents the hierarchy of targets based on whether the models answered correctly or incorrectly. 
Since $t_0$ and $t_1$ are equally informative, $t_0$ is omitted, and $t_1$ includes $t_0$.

\begin{center}
\includegraphics[width=5cm, height=6cm]{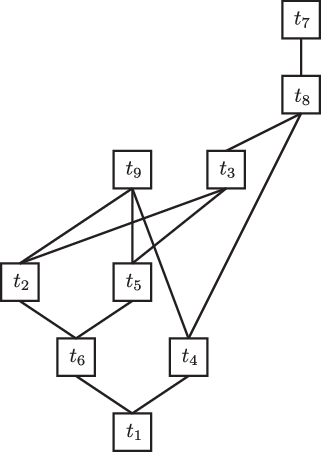}
\captionof{figure}{hierarchy of targets}
\end{center}

Figure 2 illustrates the hierarchy among judgement targets ($t_1$ to $t_9$) as perceived by models.
For instance, when $t_5$ is correctly judged, there is a tendency for $t_6$ and its subordinates, such as $t_1$, to also be correctly judged. 
Equivalently, if $t_6$ is incorrectly judged, there 
may be a superior target of $t_5$ making an incorrect judgment. This depicts a hierarchical relationship, indicating that, for example, to make accurate judgments about $t_5$, a model that correctly evaluates $t_6$ and onwards is necessary.

\bigskip

\section{Method of representing hierarchy}

\def\Tab{\text{Tab}}
\def\Ord{\text{Ord}}
\def\Has{\text{Has}}

Consider $\{t_j \}_{j=1}^u$ as the judgement targets and $\{M_i \}_{i=1}^v$  as a series of machines. Also, denote $\alpha$ as flexibility, where $\alpha$ is a positive value less than 50. In this scenario, we can succinctly express our method as follows.

\begin{enumerate}
    \item $\Tab(i,j)=
\begin{cases}
1 \,\,\, \text{(if $M_i$ judged $t_j$ correctly)}\\
0 \,\,\, \text{(if $M_i$ judged $t_j$ incorrectly)}
\end{cases}$

\item Applying the function $\Tab$ to Table 1 yields Table 4. we redefine the order $t_p \rightarrow t_q$ using $\alpha$ instead of the flexibility $m$ according to Definition \ref{ordering}. Under this definition, if $t_p \rightarrow t_q$ and $t_q \rightarrow t_p$, we can consider $t_p$ and $t_q$ as equivalent. Based on these equivalence relations, we can reclassify the set $\{ t_j \}_{j=1}^u$. To avoid cumbersome subscripts, let's proceed by classifying the results into equivalence classes, denoted as the set $\{ t_j \}_{j=1}^u$. This step is a reduction of targets based on equivalence relations.

\renewcommand{\arraystretch}{1.5}
\begin{table}[ht]
\centering
\caption{The number of machines for the judgments of $t_p$, $t_q$}
\begin{center}
\begin{tabular}[centerline]{|c|c|c|} \hline
$t_p$& $t_q$ & the number of machines  \\  \hline
1  & 1   &  $\sum_{j=1}^v \Tab(j,p )  \Tab(j,q )$ \\ \hline
1  & 0   &  $\sum_{j=1}^v  \Tab(j,p ) \left( 1- \Tab(j,q ) \right)$ \\ \hline
0  & 1   &  $\sum_{j=1}^v \left(1- \Tab(j,p )\right)  \Tab(j,q )$ \\ \hline
0  & 0   &  $\sum_{j=1}^v \left(1- \Tab(j,p ) \right) \left(1-  \Tab(j,q )\right)$ \\ \hline
\end{tabular}
\end{center}
\end{table}
\renewcommand{\arraystretch}{1}
   \item $\Ord(p,q)=
   \begin{cases}
   1 \,\,\, (t_p \rightarrow t_q ) \\
   0 \,\,\, \text{(otherwise)}
   \end{cases}$
   \item $\Has(p,r)=
   \begin{cases}
   0 \,\,\, \text{(if there is $q$ that } \Ord(p,q)=1\text{ and }\Ord(q,r)=1 \text{ hold)} \\
   \Ord(p,q)\,\,\, \text{(otherwise)}
   \end{cases}$
   \item $\Has(p,q)=1$, it is drawn as $t_p \rightarrow t_q$
\end{enumerate}

Repeating the process in step 5 iteratively allows us to visualize the hierarchy.

%%%%%%%%%%%% 참고문 %%%%%%%%%%%%%

\bibliographystyle{unsrt}

%%%%%%%%%%%%%%%%%%%%%%%%%%%%%%%%%%

\end{document}